\pdfoutput=1
\documentclass[11pt]{article}

\usepackage{acl}

\usepackage{times}
\usepackage{latexsym}

\usepackage[T1]{fontenc}
\usepackage[utf8]{inputenc}

\usepackage{microtype}

\usepackage{amsmath}
\usepackage{amssymb}
\usepackage{bm}
\usepackage{algorithm}
\usepackage[noend]{algpseudocode}
\usepackage{multirow}
\usepackage{booktabs} 
\usepackage{tabularx}
\usepackage{graphicx} 

\definecolor{yellow}{RGB}{246, 193, 192}
\definecolor{pink}{RGB}{183, 178, 254}
\definecolor{cyan}{RGB}{205, 207, 196}

\usepackage{transparent}

\title{MemSum: Extractive Summarization of Long Documents Using \\ Multi-Step Episodic Markov Decision Processes}

\author{Nianlong Gu \\
  Institute of Neuroinformatics,\\ 
  University of Zurich and\\
  ETH Zurich\\
  \texttt{nianlong@ini.ethz.ch} \\\And
  Elliott Ash \\
  Department of Humanities,\\
  Social and Political Sciences, \\
  ETH Zurich \\
  \texttt{ashe@ethz.ch} \\ \And
  Richard H.R. Hahnloser\\
  Institute of Neuroinformatics,\\
  University of Zurich and\\
  ETH Zurich\\
  \texttt{rich@ini.ethz.ch}\\
  }

\begin{document}
\maketitle
\begin{abstract}
We introduce MemSum (Multi-step Episodic Markov decision process extractive SUMmarizer), a reinforcement-learning-based extractive summarizer enriched at each step with information on the current extraction history. When MemSum iteratively selects sentences into the summary, it considers a broad information set that would intuitively also be used by humans in this task: 1) the text content of the sentence, 2) the global text context of the rest of the document, and 3) the extraction history consisting of the set of sentences that have already been extracted. With a lightweight architecture, MemSum obtains state-of-the-art test-set performance (ROUGE) in summarizing long documents taken from PubMed, arXiv, and GovReport. Ablation studies demonstrate the importance of local, global, and history information. A human evaluation confirms the high quality and low redundancy of the generated summaries, stemming from MemSum's awareness of extraction history.
\end{abstract}

\section{Introduction}

Automatic text summarization is the task of automatically summarizing a long document into a relatively short text while preserving most of the information \cite{tas2007survey}. Text summarization methods can be categorized into abstractive and extractive summarization \cite{10.1007/s10462-016-9475-9,nenkova_survey_2012}. Given a document $d$ consisting of an ordered list of $N$ sentences, \textit{extractive summarization} aims to pick up $M$ ($M$$\ll$$N$) sentences as the summary of the document. The extracted summaries tend to be both grammatically and semantically more reliable than abstractive summaries \cite{liu2018generating, liu2019hierarchical, luo2019reading,liao2020improving}, as they are directly selected from the source text.

Extractive summarization is usually modeled as two sequential phases \cite{zhou-etal-2018-neural-document}: 1) \textit{sentence scoring} and 2) \textit{sentence selection}. In the sentence scoring phase, an affinity score is computed for each sentence by neural networks such as bidirectional RNNs \cite{dong2018banditsum,narayan2018ranking,luo2019reading,xiao-carenini-2019-extractive} or BERT \cite{zhang2019hibert,liu2019text}. In the sentence selection phase, sentences are selected by either i) predicting a label (1 or 0) for each sentence based on its score, and selecting sentences with label 1 \cite{zhang2019hibert,liu2019text,xiao-carenini-2019-extractive}, or ii) ranking sentences based on their scores and selecting the top $K$ sentences as the summary \cite{narayan2018ranking}, or iii) sequentially sampling sentences without replacement, where the normalized scores of the remaining sentences are used as sampling likelihoods \cite{dong2018banditsum,luo2019reading}.
%Typically, the sentence scores keep unchanged during the sentence selection phrase, and sentences with higher scores are more likely to be selected into the summary. 

In these approaches, sentence scores are generally not updated based on the current partial summary of previously selected sentences, indicating a lack of knowledge of \textit{extraction history}. We deem extractive summarizers that are not aware of the extraction history to be susceptible to redundancy in a document, because they will repeatedly add sentences with high scores to a summary, regardless of whether similar sentences have been selected before. And, redundancy leads to performance decreases evaluated by ROUGE F1.

\begin{figure}
\centering
  \includegraphics[width=\linewidth]{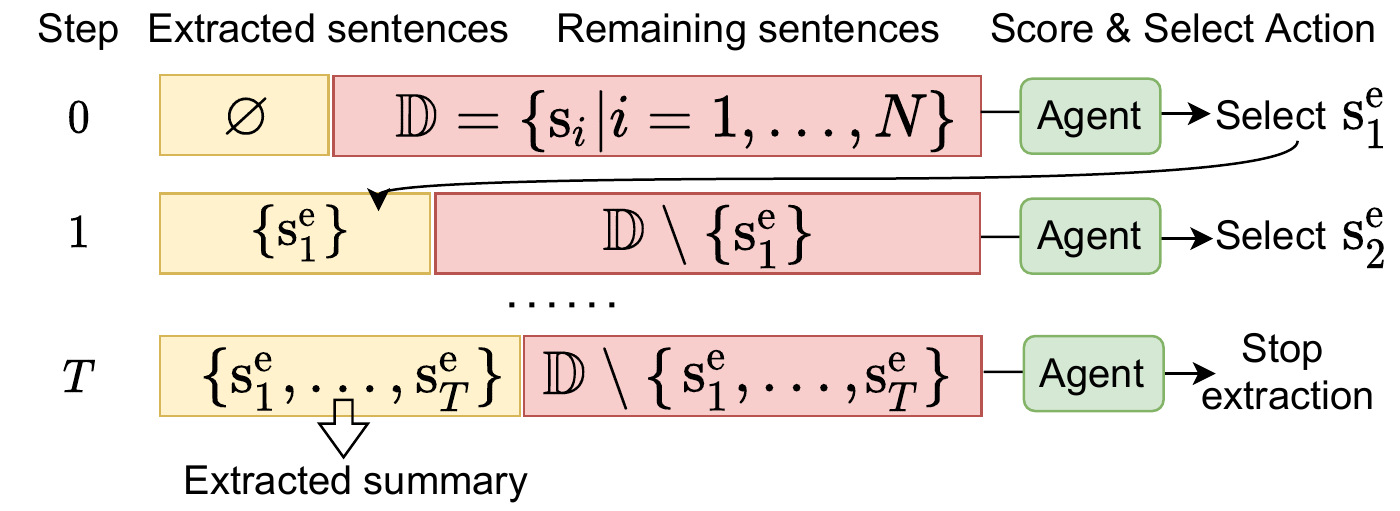}
  \caption{We modeled extractive summarization as a multi-step iterative process of scoring and selecting sentences. $\text{s}_i$ represents the $i_\text{th}$ sentence in the document $\mathbb{D}$.}
  \label{fig:extractive_summ_pipeline}
\end{figure}

In this paper, we propose to model extractive summarization as a multi-step episodic Markov Decision Process (MDP). As shown in Figure \ref{fig:extractive_summ_pipeline}, at each time step in an episode, we define a \textit{sentence state} composed of three sub-states: 1) the local content of the sentence, 2) the global context of the sentence within the document, and 3) information on the extraction history, including the previously selected set of unordered sentences and the remaining sentences. At each time step, the policy network (agent) takes the current sentence state as input and produces scores used to select an action of either stopping the extraction process or selecting one of the remaining sentences into the candidate summary. Unlike one-step episodic MDP-based models \cite{narayan2018ranking,dong2018banditsum, luo2019reading} that encode the state information only once at the beginning of the episode, in our multi-step policy, the agent updates at each time step the extraction history before selecting an action. Such a step-wise state-updating strategy enables the agent to consider the content of the partial summary when selecting a sentence. 

To efficiently encode local and global sentence states, we design an extraction agent based on LSTM networks \cite{hochreiter1997long}. To encode the extraction history and to select actions, we use a reduced number of attention layers \cite{vaswani2017attention} of relatively low dimensionality. These choices enable our model to be easily trainable and to summarize long documents such as scientific papers \cite{cohan2018discourse,huang2021efficient} or reports \cite{huang2021efficient}.

The \textbf{contributions} of our work are as follows: 1) We propose to treat extractive summarization as a multi-step episodic MDP that is aware of the extraction history.
2) We show that extraction-history awareness allows our model to extract more compact summaries than models without history awareness and behave more robustly to redundancies in documents. 3) Our model outperforms both extractive and abstractive summarization models on PubMed, arXiv \cite{cohan2018discourse}, and GovReport \cite{huang2021efficient} datasets. 4) Finally, human evaluators rate the MemSum summaries to be of higher quality than those from a competitive approach, especially by virtue of lower redundancy\footnote{Our code and data are available at \url{https://github.com/nianlonggu/MemSum}}.

\section{Related Work}
\label{sec:related_works}

Extraction history awareness was previously considered in NeuSum \cite{zhou-etal-2018-neural-document}, where a GRU encoded previously selected sentences into a hidden vector that then was used to update the scores of the remaining sentences to bias the next selection. NeuSum contains no stopping mechanism and therefore it can only extract a fixed number of sentences, which likely is sub-optimal. Also, the potential benefits of extraction history have not been quantified and so the idea remains unexplored to a large extent.
% Another way to make the extraction process history-aware is to select actions by considering a combination of multiple sentences as the candidate summary. For example, a two-stage extractive summarizer is provided in MatchSum \cite{zhong2020extractive}: First, a BERT-based summarizer \cite{liu2019text} is used to obtain a pruned subset of salient sentences. Second, all combinations of sentences from that pruned subset are ranked by a fine-tuned Siamese-BERT \cite{devlin2018bert}.
% based on the assumption that better candidate summaries should be semantically closer to the original document. 

Recently, BERT-based extractors such as MatchSum \cite{zhong2020extractive} achieved SOTA performance in extractive summarization of relatively short documents from the CNN/DM \cite{hermann2015teaching} dataset. However, the quadratic computational and memory complexities \cite{huang2021efficient} of such models limit their scalability for summarizing long documents with thousands of tokens, which is common for scientific papers and government reports. Although large pre-trained transformers with efficient attention \cite{huang2021efficient} have been adapted for abstractive summarization of long documents, we believe that extractive summarization is more faithful in general, which is why we chose an extractive approach.

% but they are limited in several ways. In particular, BERT-based extractors inherit BERT's sequence length limitation of 512 tokens \cite{devlin2018bert}, precluding extractive summarization of long documents. In addition, BERT-based extractors contain numerous trainable parameters and so fine-tuning them requires a large corpus of document-summary pairs.

\section{Model}
\begin{figure*}
\centering
  \includegraphics[width=\linewidth]{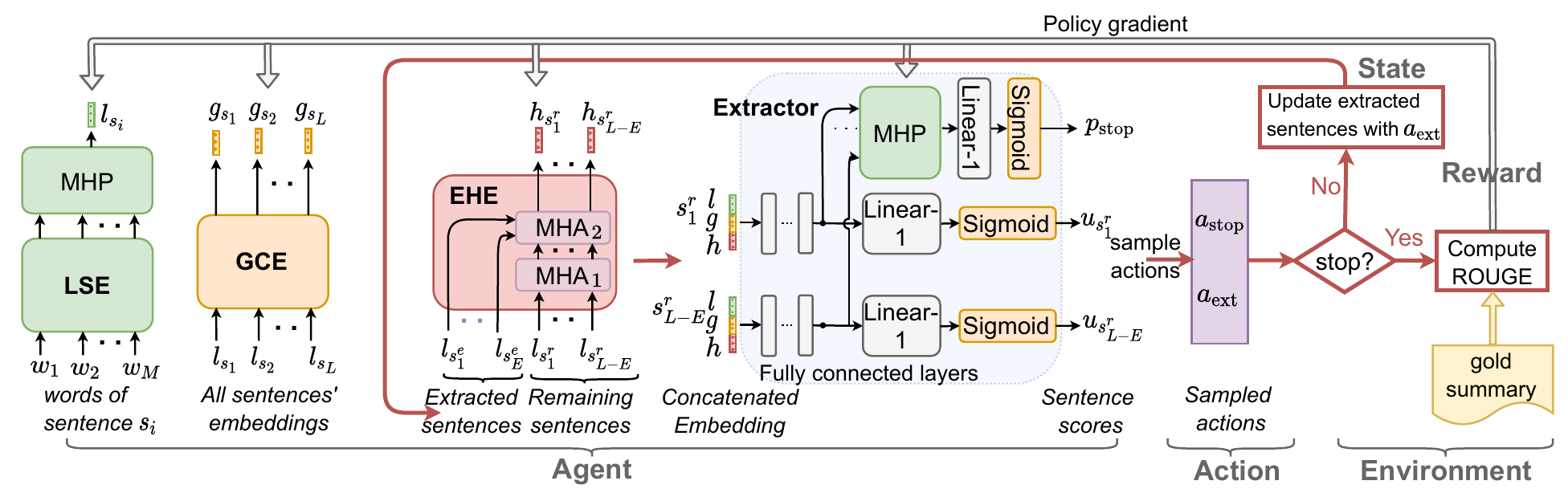}
  \caption{The architecture of our MemSum extractive summarizer with a multi-step episodic MDP policy.  With the updating of the extraction-history embeddings $h$ at each time step $t$,  the scores $u$ of remaining sentences and the stopping probability $p_\text{stop}$ are updated as well.
  }
  \label{fig:model_architecture}
\end{figure*}
This section outlines the multi-step episodic MDP policy for extractive summarization. 
\subsection{Policy Gradient Methods}
% Policy gradient methods aim to learn a parameterized policy of selecting actions that maximize a certain objective function $J(\bm{\theta})$, where $\bm{\theta}$ represents the parameters of the policy $\pi_{\bm{\theta}}$. In an episodic task with a terminal state (i.e. \textit{end of summary}), the objective function $J$ is the expected return for that policy, $J(\bm{\theta})=\mathbb{E}_{\pi_{\bm{\theta}}}[R_0]$, where the return $R_t=\sum_{k=t+1}^{T}r_k$ is the cumulative reward from time $t+1$ until the end of the episode when the summary is complete. In applications of RL to extractive summarization, the instantaneous reward $r_t$ is zero except at the end of the episode when the final reward $r$ is computed according to Equation \eqref{eq:R_compute}, so $R_t\equiv R_0=r$. The reward $r$ is usually expressed as

% Policy gradient methods aim to learn a parameterized policy of selecting actions that maximize a certain objective function $J(\bm{\theta})$, where $\bm{\theta}$ represents the parameters of the policy $\pi_{\bm{\theta}}$. 

In an episodic task with a terminal state (i.e. \textit{end of summary}), policy gradient methods aim to maximize the objective function  $J(\bm{\theta})=\mathbb{E}_{\pi_{\bm{\theta}}}[R_0]$, where the return $R_t=\sum_{k=t+1}^{T}r_k$ is the cumulative reward from time $t+1$ until the end of the episode when the summary is complete. In applications of RL to extractive summarization, the instantaneous reward $r_t$ is zero except at the end of the episode when the final reward $r$ is computed according to Equation \eqref{eq:R_compute}, so $R_t\equiv R_0=r$. The reward $r$ is usually expressed as
\cite{dong2018banditsum}:
\begin{equation}
\label{eq:R_compute}
r = \frac{1}{3}( \text{ROUGE-1}_f +\text{ROUGE-2}_f+\text{ROUGE-L}_f  )
\end{equation}
According to the REINFORCE algorithm \cite{williams1992simple}, the policy gradient is defined as:
\begin{equation}
    \nabla J(\bm{\theta}) = \mathbb{E}_\pi[R_t\nabla\log\pi(A_t\vert S_t,\bm{\theta})],
\end{equation}
where $\pi(A_t\vert S_t, \bm{\theta})$ denotes the likelihood that at time step $t$ the policy  $\pi_{\bm{\theta}}$ selects action $A_t$ given the state $S_t$.
With $\alpha$ as the learning rate, the parameter update rule is \cite{sutton2018}:
\begin{equation}
    \label{eq:update_rule}
    \bm{\theta}_{t+1} \leftarrow \bm{\theta}_t + \alpha R_t \nabla\log\pi(A_t\vert S_t,\bm{\theta}_t),%=\alpha R \nabla\log p_{\bm{\theta}}(a\vert S_0)
\end{equation}
% where $\alpha$ is the learning rate.

\subsection{Multi-step Episodic MDP Policy}

% The innovation of our model is to take an episodic approach. 
% We define an action as the selection of a single sentence and an episode as the generation of a summary. 
% The objective, then, is to learn a policy that maximizes the expected return of an episode. 
Different from one-step episodic MDP policies \cite{narayan2018ranking,dong2018banditsum,luo2019reading} that extract the entire summary via a single action, we define an episode, i.e., the generation of a summary, consisting of multiple time steps. At each time step $t$, corresponding to extracting sentence number $t$, the action $A_t$ is either to stop extraction or to select a sentence $s_{a_t}$ from the remaining sentences. The agent's policy is:
\begin{equation}
\label{eq:memsum_action_prob}
\begin{aligned}
\pi(A_t\vert S_t,\bm{\theta}_t) &= p(\text{stop}|S_t, \bm{\theta}_t) p(a_t|\text{stop},S_t, \bm{\theta}_t)\\
p(a_t|\text{stop},S_t, \bm{\theta}_t) &= \begin{cases}
\frac{ u_{a_t}(S_t,\bm{\theta}_t) }{ \sum_{j\in I_t} u_j( S_t,\bm{\theta}_t ) } \ &\text{if} \ \text{stop = false} \\
\frac{1}{\vert I_t \vert} &\text{if} \ \text{stop = true},
\end{cases}
\end{aligned}
\end{equation}
where $I_t$ denotes the index set of remaining sentences at time step $t$. If the agent does not stop, it first computes a score $u_j$ for each remaining sentence and samples a sentence $s_{a_t}$ according to the probability distribution of normalized scores. 
% Sentence $s_{a_t}$ is selected with conditional likelihood $p(a_t|\text{stop=false},S_t, \bm{\theta}_t)$. 
When the agent stops the extraction, no sentence is selected and the conditional likelihood $p(a_t|\text{stop=false},S_t, \bm{\theta}_t)$ is set to $\frac{1}{\vert I_t \vert}$ (where $\vert I_t \vert$ represents the number of remaining sentences at time $t$), which is independent of the policy parameters to prohibit the gradient from being passed to the policy parameters via the conditional likelihood. After calculating the reward according to Equation~\eqref{eq:R_compute}, the policy parameters are updated according to Equation~\eqref{eq:update_rule} (for all time steps).

\subsection{Policy Network}
\label{sec:policy_network_structure}
The state $S_t$ in Equation \eqref{eq:memsum_action_prob} is designed to be informative on: 1) the local content of the sentence, 2) the global context of the sentence within the document, and 3) the current extraction history. To encode these three properties in the state, we use a local sentence encoder, a global context encoder, and an extraction history encoder, respectively. Subsequently, the state is mapped by an extractor to an output score for each of the remaining sentences and the extraction stop signal. The overall framework of our model is depicted in Figure~\ref{fig:model_architecture}.

% \textbf{Local Sentence Encoder}
In the \textbf{Local Sentence Encoder} (LSE), ordered words $(w_1, w_2, \dots w_M)$ in a sentence $s_i$ are first mapped onto word embeddings using a word embedding matrix. Subsequently, a $N_l$-layer bi-directional LSTM \cite{hochreiter1997long} transforms the word embeddings and maps them onto sentence embeddings $l_{s_i}$ via a multi-head pooling layer (MHP)~\cite{liu2019hierarchical}.

% \textbf{Global Context Encoder}
The \textbf{Global Context Encoder} (GCE) consists of a $N_g$-layer bi-LSTM that takes the $L$ local sentence embeddings $(l_{s_1}, l_{s_2}, \dots l_{s_L})$ as inputs and produces for each sentence $s_i$ an embedding $g_{s_i}$ that encodes global contextual information such as the sentence's position in the document and information on neighboring sentences.

The \textbf{Extraction History Encoder} (EHE) encodes the extraction history information and produces the extraction history embedding $h_{s^r_i}$ for each remaining sentence $s^r_i$.
The EHE is composed of a stack of $N_h$ identical layers. Within one layer, there are two multi-head attention sublayers, as contained in the transformer decoder in \citet{vaswani2017attention}. One  sublayer is used to perform multi-head self-attention (MHA) among the local embeddings of the remaining sentences, so that each remaining sentence can capture the context provided by other remaining sentences. The other attention sublayer is used to perform multi-head attention over the embeddings of extracted sentences to enable each remaining sentence to attend to all the extracted sentences. The output of the two attention sublayers, one for each remaining sentence, captures the contextual information of both extracted and remaining sentences. The final output of the $N_h^{\text{th}}$ layer of the EHE constitutes the extraction history embedding, one for each remaining sentence. 

There is no positional encoding and the EHE produces the extraction history embeddings non-autoregressively by attending to both precedent and subsequent positions. Consequently, the extraction history embeddings $h_{s^r_i}$  for the remaining sentences are invariant to the order of the previously selected sentences. We believe that the sequential information of previously selected sentences is not crucial for reducing redundancy and for deciding whether to stop extraction or not.

The \textbf{Extractor} computes the score of each remaining sentence and outputs an extraction stop signal. As input to the extractor, we form for each of the remaining sentences $s^r_i$ an aggregated embedding by concatenating the local sentence embedding $l_{s^r_i}$, the global context embedding $g_{s^r_i}$, and the extraction history embedding $h_{s^r_i}$. As shown in Figure \ref{fig:model_architecture}, to produce the score $u_{s^r_i}$, the concatenated embedding of remaining sentence $s^r_i$ is passed to fully connected layers with ReLU activation and then projected to a scalar by a Linear-1 layer followed by a sigmoid function. Note that the same fully connected layers are applied identically to all remaining sentences. We deem that the extractor can learn to stop extraction based on the remaining sentences' states. Therefore, we apply an MHP to the last hidden vectors of all remaining sentences to output a single vector. This vector is then passed to a linear layer with a sigmoid function, producing a stopping probability~$p_\text{stop}$.
% the extractor is a fully connected network in which a multi-head pooling layer \cite{liu2019hierarchical} maps the last hidden vectors of all remaining sentences into a single vector. This vector is then passed to a linear layer with a sigmoid activation function, producing a stopping probability $p_\text{stop}$ as well as a score $u_{s_i^r}$ for each. 
% $u_{s_i^r}$ can be interpreted as the affinity of being extracted into the candidate summary \cite{dong2018banditsum} at the current time step. 

\subsection{Training}
\label{sec:training}
We train the parameterized policy network according to the update rule in Equation \eqref{eq:update_rule}. At each training iteration, an episode is sampled to compute the final return $r$ and the action probabilities $\pi(A_t\vert S_t, \bm{\theta}_t)$ for all time steps $t$. An example episode with $T$ extracted sentences looks like: $(S_0, s_{a_0},\dots,S_{T-1},s_{a_{T-1}}, S_{T}, A_{\text{stop}},r)$, where $S_t$ represents the concatenated state information introduced in Section \ref{sec:policy_network_structure}, $s_{a_t}$ represents the selection of sentence $a_t$, $A_{\text{stop}}$ represents the extraction stops at the final time step $T$, and $r$ is the reward as defined in Equation~\eqref{eq:R_compute}. To encourage the agent to select compact summaries, we multiply the final reward $r$ by a length penalty term $1/(T+1)$  \cite{luo2019reading}. Consequently, the return $R_t\equiv\frac{r}{T+1}$.
\begin{algorithm}
\caption{The training algorithm.}
\label{alg:training}
Parameters: learning rate $\alpha$
\begin{algorithmic}[1]
\For{each document-summary pair ($D_i$, $G_i$)}
    \State LSE outputs local sent. embed $l_{s_1}$,$\dots$,$l_{s_L}$
    \State GCE outputs global context embed $g_{s_1}$,$\dots$,$g_{s_L}$
    \State Sample an episode $S_0$,$s_{a_0}$,$\dots$,$S_{T-1}$,$s_{a_{T-1}}$, $S_{T}$,$A_{\text{stop}}$,$r$ from the high-ROUGE episodes set $\mathbb{E}_p$ of document $D_i$ 
    % \State Randomly shuffle the order of selected sentences $s_{a_0}$,...,$s_{a_{T-1}}$ and update the states $S_0$,...,$S_{T-1}$
    \For{each time step: $t$ = 0,1,...,T:}
        \If{$t>0$}
            \State EHE outputs extraction history embed $h_{s^r_1}$,$\dots$,$h_{s^r_{L-E_{t}}}$ for remaining sentences
        \Else
               \State Initialize $h_{s^r_1}$,...,$h_{s^r_{L-E_{0}}}$ to $\bm{0}$
        \EndIf
        \State Extractor outputs scores $u_{s_1^r}$,...,$u_{s_{L-E_t}^r}$ for remaining sentences and outputs $p_{\text{stop}}$
        \State Compute the action probability $\pi(A_t\vert S_t,\bm{\theta})$ according to Equation \eqref{eq:memsum_action_prob}
        \State $\bm{\theta} \leftarrow \bm{\theta} + \alpha \frac{r}{T+1} \nabla \log \pi(A_t\vert S_t,\bm{\theta}) $
    \EndFor
\EndFor
\end{algorithmic}
\end{algorithm}

Algorithm \ref{alg:training} summarizes the training procedure of MemSum. 
% We note two implementation details. First, in line 5, we randomly shuffle the order of the selected sentences. Because the extraction order does not influence the ROUGE score, this shuffling operation can prevent the agent from overfitting the sentence selection order. Second, 
We initialize the extraction history embeddings to $\bm{0}$, because at $t=0$ no sentences have been extracted. $E_t$ represents the number of sentences that have been extracted into the summary up to time step~$t$. Following the strategy in \citet{narayan2018ranking} and \citet{mohsen2020hierarchical}, instead of sampling an episode following the current policy $\pi(\cdot\vert \cdot,\bm{\theta}_t)$, we sample an episode from a set $\mathbb{E}_p$ of episodes with high ROUGE scores, which enables the agent to quickly learn from optimal policies and to rapidly converge. Details on creating a set of high-ROUGE episodes for training are described in Appendix \ref{sec:high_rouge}.

\section{Experiments}
\label{sec:experiments}
In this section, we report implementation details of our model and describe the datasets used for training and for evaluation.

\noindent \textbf{Datasets.}
The documents to be summarized in the PubMed and arXiv datasets \cite{cohan2018discourse} are the full bodies of scientific papers and the gold summaries are the corresponding abstracts. \citet{zhong2020extractive} proposed a truncated version of the PubMed dataset (PubMed$_\text{trunc}$ for simplicity) by defining a doument as the introduction section of a paper. The GovReport dataset \cite{huang2021efficient} contains U.S. government reports with gold summaries written by experts.
% Table \ref{tab:datasets} reports for each dataset the average document and summary length, as well as the number of document-summary pairs for training, validating, and testing. 
Except PubMed$_\text{trunc}$, all the other datasets contain significantly longer documents than the popular dataset CNN/DM (Table~\ref{tab:datasets}).

% We simply treat the document as a list of sentences without considering section information, following a general setting adopted in recent works \cite{dong2018banditsum,zhong2020extractive} but unlike \citet{ xiao-carenini-2019-extractive,huang2021efficient}, 

\begin{table}

\centering
\resizebox{\linewidth}{!}{ 
\setlength\tabcolsep{2.5 pt}
\begin{tabular}{cccccccccc}
\toprule 
 \multirow{3}*{\textbf{Datasets}} &  \multicolumn{2}{c}{  \shortstack{ \textbf{avg. doc.}\\\textbf{length} } } & & \multicolumn{2}{c}{ \shortstack{ \textbf{avg. summ.}\\\textbf{length} } } & & \multicolumn{3}{c}{\shortstack{\textbf{\# of doc.-summ.}\\\textbf{pairs}}}\\

 \cmidrule(lr){2-3}
\cmidrule(lr){5-6}
\cmidrule(lr){8-10}
 &  \multirow{2}*{\shortstack{\# of\\words}} & \multirow{2}*{\shortstack{\# of\\sent.}}& &  \multirow{2}*{\shortstack{\# of\\words}} & \multirow{2}*{\shortstack{\# of\\sent.}} & & \multirow{2}*{Train} & \multirow{2}*{Valid} & \multirow{2}*{Test} 
 \\ \\ \midrule
 PubMed & 2,730 & 88 & & 181  & 7 & &  116,937 & 6,633 & 6,658 \\
 arXiv &5,206 & 206 & & 238  & 10 &  & 202,880 & 6,436  & 6,440 \\
 PubMed$_\text{trunc}$ & 408 & 13 & & 185  & 7  && 83,233  & 4,676 & 5,025 \\
 GovReport & 7,932 & 307 & & 501 & 18  & & 17,517 & 974 & 973 \\
  CNN/DM & 692 & 35 & & 49 & 4  & & - & - & - \\
\bottomrule
\end{tabular}
}
\caption{ \label{tab:datasets} An overview of datasets used in this paper. We count only strings composed of letters and numbers for \# of words. }
\end{table}
\noindent \textbf{Baselines.}
% We compared our MemSum with both extractive and abstractive models.
Extractive baselines include Lead (directly using the first several sentences as the summary) \cite{dancerp}, SummaRuNNer \cite{nallapati2016summarunner}, Atten-Cont \cite{xiao-carenini-2019-extractive}, Sent-CLF and Sent-PTR \cite{ pilault-etal-2020-extractive}, MatchSum \cite{zhong2020extractive}, and the NeuSum model \cite{zhou-etal-2018-neural-document} that we trained on our datasets.

Abstractive summarization models include PEGASUS \cite{pmlr-v119-zhang20ae}, BigBird \cite{NEURIPS2020_c8512d14}, Dancer \cite{dancerp}, and Hepos \cite{huang2021efficient} that achieved the state-of-the-art in long document summarization using a large-scale pretrained BART model \cite{lewis-etal-2020-bart} with memory-efficient attention encoding schemes including Locality Sensitive Hashing \cite{Kitaev2020Reformer:} (Hepos-LSH) and Sinkhorn attention (Hepos-Sinkhorn). We also present the performance of the oracle extraction model based on the greedy approach \cite{nallapati2016summarunner} which sequentially selects from the document the sentence that maximally improves the average of R-1 and R-2 of selected sentences. 
% We also evaluated the performance of our own Lead-10 baseline on PubMed and arXiv datasets.

\noindent\textbf{Implementation Details.} 
We computed local sentence embeddings using pretrained Glove word embeddings \cite{pennington2014glove} of dimension $d=200$, keeping the word embeddings fixed during training. For the LSE, we used $N_l=2$ bi-LSTM layers and for the GCE $N_g=2$. For the EHE, we used $N_h=3$ attention layers, and we set the number of attention heads to $8$ and the dimension of the feed-forward hidden layer to $1024$; during training we set the dropout rate to $0.1$. The extractor consisted of $2$ fully-connected hidden layers with output dimensions $2d$ and $d$, respectively. 
% The total number of trainable parameters is 4.4 M, so our model is much smaller than BERT-based extractive summarization models in which merely the $\text{BERT}_\text{BASE}$ \cite{devlin2018bert} contains 110 M trainable parameters. 

We trained our model using the Adam optimizer with $\beta_1=0.9$, $\beta_2=0.999$ \cite{DBLP:journals/corr/KingmaB14}, fixed learning rate $\alpha=1e^{-4}$, and weight decay $1e^{-6}$. 
% To store the network parameters for validating and inference, we used the exponential moving average strategy introduced in \citet{karras2017progressive} with decay $0.999$. 
The training was stopped when the validation performance started to degrade.
% After the model parameters $m$ being updated after each training iteration, we update the exponential moving average parameters $m_\text{ema}$ according to $m_\text{ema}\leftarrow \beta m_\text{ema} + (1-\beta) m$. 
During validating and testing, the agent extracted sentences in a deterministic way: after computing the scores $u_{s_i^r}$ for the remaining sentences and the stop likelihood $p_\text{stop}$, the agent stopped the extraction if $p_\text{stop}\geq p_\text{thres}$ or if the maximum admissible number $N_\text{max}$ of extracted sentences was reached; otherwise, the agent selected the sentence with the largest score. The model was trained on eight RTX 2080 Ti GPUs. 

On the validating datasets we selected the best checkpoint of each model and determined the optimal $N_\text{max}$ and stopping criterion $p^*_\text{thres}$. For Pubmed, arXiv, Pubmed$_\text{trunc}$, and GovReport, $N_\text{max}$ was set to $7$, $5$, $7$, and $22$, and  $p^*_\text{thres}$ was set to $0.6$, $0.5$, $0.8$, and $0.6$, respectively. For the detailed selection procedure of the optimal stopping threshold, see Appendix \ref{sec:selection_of_optimal}. Information on reproducibility is available in Appendix \ref{sec:reproduce}.

\noindent\textbf{Evaluation.} We evaluated the performance of our model using $F_1$ ROUGE \cite{lin2004rouge}, including ROUGE-1,2, and L for measuring unigram, bigram, and longest common subsequence. We also conducted human evaluation in Section \ref{sec:human-evaluation}.
%We used a native python implementation of ROUGE for training and validating and used the official pyrouge package\footnote{We use pyrouge, a python wrapper of ROUGE, with the parameters "-a -c 95 -m -n 4 -w 1.2".} for testing  \cite{luo2019reading,dong2018banditsum}.

\section{Results and Discussion}
Here we present the results on various extractive summarization tasks and analyze the contribution of different modules via ablation studies.
\subsection{Results Comparison}

% and obtained ROUGE scores close to the ones provided in \cite{dancerp}, which indicates that  our test dataset and ROUGE computation process is consistent with previous works.

\begin{table}
\centering
\resizebox{\linewidth}{!}{ 
\begin{tabular}{lcccccc}
\toprule 
\multirow{2}*{\textbf{Model}}  & \multicolumn{3}{c}{\textbf{PubMed}} & \multicolumn{3}{c}{\textbf{arXiv}} \\
\cmidrule(lr){2-4}
\cmidrule(lr){5-7}
  & R-1  &  R-2   &  R-L & R-1  &  R-2   &  R-L \\
  \hline \\[-1em]
ORACLE & 61.99  & 34.95 & 56.76 & 60.00  & 30.60 & 53.03  \\ \hline \\[-1em]
\multicolumn{4}{l}{\textbf{Extractive summarization baselines}}
\\ \\[-1em]
% SumBasic & 37.15 & 11.36 & 33.43 & 29.47 & 6.95 & 26.3 \\
Lead-10  & 37.45 & 14.19 & 34.07  & 35.52 & 10.33 & 31.44\\
SummaRuNNer   & 43.89 & 18.78 & 30.36 & 42.81 & 16.52 & 28.23\\
Atten-Cont  & 44.85 & 19.70 & 31.43  & 43.62 & 17.36 & 29.14 \\
Sent-CLF   & 45.01  & 19.91  & 41.16  & 34.01  & 8.71  & 30.41\\ 
Sent-PTR   & 43.30  & 17.92  & 39.47  & 42.32  & 15.63  & 38.06\\ 
NeuSum   &  47.46 & 21.92  & 42.87  & 47.49   &21.56  & 41.58 \\
\hline
\\[-1em]
\multicolumn{4}{l}{\textbf{Abstractive summarization baselines}}
\\ \\[-1em]
PEGASUS   & 45.97 & 20.15 & 41.34 & 44.21 & 16.95 & 38.83 \\
BigBird   & 46.32 & 20.65 & 42.33 & 46.63 & 19.02 & 41.77 \\
Dancer & 46.34 & 19.97 & 42.42 & 45.01 & 17.60 & 40.56\\
Hepos-Sinkhorn  & 47.93 & 20.74 &42.58 & 47.87 & 20.00 &41.50\\
Hepos-LSH  & 48.12 & 21.06 & 42.72  & 48.24 & 20.26 & 41.78\\ \hline
\\[-1em]
% \multicolumn{4}{l}{\textbf{Our Models}}
% \\ \\[-1em]
% Lead-10 (ours) & 37.71  & 14.13  & 34.30 & 34.88  & 10.46  & 30.96 \\
\textbf{MemSum (ours)}  & \textbf{49.25}* & \textbf{22.94}* & \textbf{44.42}* & \textbf{48.42} & \textbf{20.30} & \textbf{42.54}*  \\

\bottomrule
\end{tabular}
}
\caption{ \label{tab:res_pubmed} Results on the PubMed and arXiv test sets. ``*" indicates that they are statistically significant in comparison to the closest baseline with a 95\% bootstrap confidence interval estimated by the ROUGE script\footnote{}. }
\end{table}

By comparing with extractive baselines on the PubMed and arXiv datasets, we observed that models utilizing extraction history, such as NeuSum and our MemSum, perform significantly better than other models, revealing the effectiveness of the extraction history. MemSum also significantly outperformed NeuSum, suggesting a \textit{better utilization of extraction history}, which we ascribed to the following factors: 1) In MemSum, we treat stopping extraction also as an action and train the policy network to output a stopping probability. Therefore, MemSum is able to automatically stop extracting at an optimal time step based on extraction history, while NeuSum can only extract a predefined number of sentences; 2) With the policy gradient method REINFORCE we can train MemSum to maximize the ROUGE score directly, while in NeuSum the loss was set to the KL-divergence between the model-computed sentence scores and the ROUGE score gains at each step, which is less intuitive. We further compare MemSum with NeuSum via human evaluation in Section \ref{sec:human-evaluation}. 

\footnotetext{\url{https://pypi.org/project/rouge-score/}}

\begin{table}
\centering
\resizebox{\linewidth}{!}{ 
\begin{tabular}{lcccccc}
\toprule 
\multirow{2}*{\textbf{Model}}  & \multicolumn{3}{c}{\textbf{PubMed$_\text{trunc}$}} & \multicolumn{3}{c}{\textbf{GovReport}} \\
\cmidrule(lr){2-4}
\cmidrule(lr){5-7}
  & R-1  &  R-2   &  R-L & R-1  &  R-2   &  R-L \\
  \hline \\[-1em]
ORACLE & 45.12  & 20.33 & 40.19 & 75.56  & 45.91 & 72.51  \\ \hline \\[-1em]
\multicolumn{4}{l}{\textbf{Extractive summarization baselines}}
\\ \\[-1em]
Lead &   37.58 & 12.22 & 33.44 & 50.94  & 19.53  & 48.45  \\
MatchSum  & 41.21 & 14.91 &  36.75  &  - & -  &- \\
NeuSum   &  - & -  & -  & 58.94 &  25.38 & 55.80 \\
\hline
\\[-1em]
\multicolumn{4}{l}{\textbf{Abstractive summarization baselines}}
\\ \\[-1em]
Hepos-LSH  & - & - & -  & 55.00 & 21.13 & 51.67\\
Hepos-Sinkhorn  & - & - &- & 56.86 & 22.62 & 53.82 \\
\hline
\\[-1em]
% \multicolumn{4}{l}{\textbf{Our Models}}
% \\ \\[-1em]
% Lead-10 (ours) & 37.71  & 14.13  & 34.30 & 34.88  & 10.46  & 30.96 \\
\textbf{MemSum (ours)} & \textbf{43.08}* & \textbf{16.71}* & \textbf{38.30}* &  \textbf{59.43}* &  \textbf{28.60}* &  \textbf{56.69}*  \\

\bottomrule
\end{tabular}
}
\caption{ \label{tab:res_gov-report} Results on PubMed$_\text{trunc}$ and GovReport. }

\end{table}

% \begin{table}
% \centering
% \resizebox{0.65\linewidth}{!}{ 
% \begin{tabular}{lccc}
% \toprule
% \textbf{Model}  & \textbf{R-1}  &   \textbf{R-2}   &  \textbf{R-L}
%  \\ \hline \\[-1em]
% ORACLE & 75.56  & 45.91 & 72.51 \\ \hline \\[-1em]
% \multicolumn{4}{l}{\textbf{Abstractive summarization baselines}}
% \\ \\[-1em]
% Hepos-LSH  & 55.00 & 21.13 & 51.67 \\ 
% Hepos-Sinkhorn & 56.86 & 22.62 & 53.82 \\ \hline
% \\[-1em]
% \multicolumn{4}{l}{\textbf{Our Models}}
% \\ \\[-1em]
% % Lead-20 (ours) & 50.94  & 19.53  & 48.45 \\
% MemSum (ours) &  \textbf{59.43}* &  \textbf{28.60}* &  \textbf{56.69}* \\

% \bottomrule
% \end{tabular}
% }
% \caption{ \label{tab:res_gov-report} Results on the GovReport test set. } 

% \end{table}

\begin{figure}[t]
\centering
  \includegraphics[width=\linewidth]{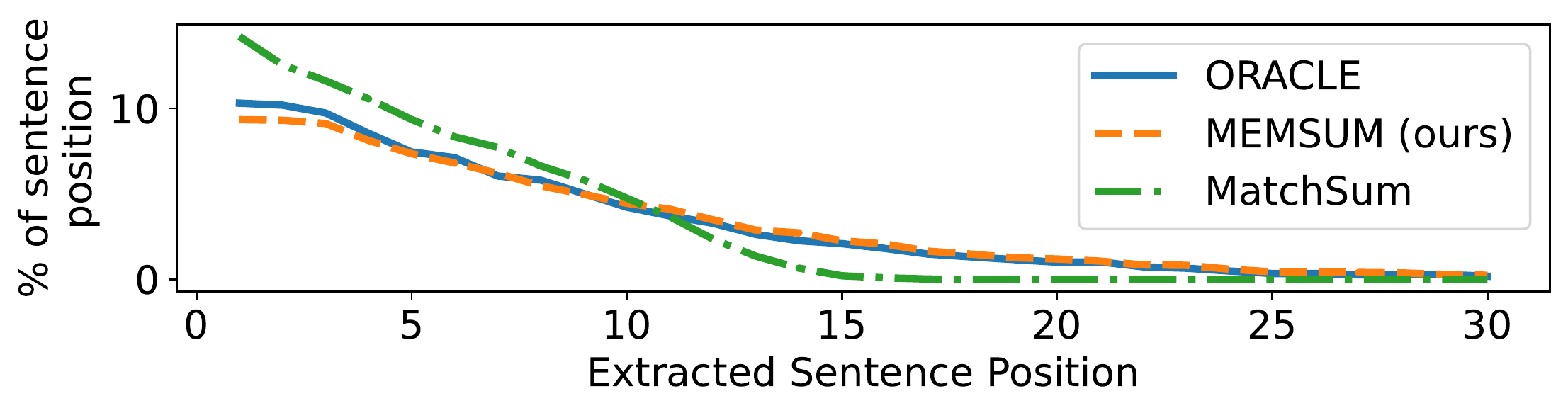}
  \caption{The position distribution of extracted sentences in the PubMed$_\text{trunc}$ dataset. }
  \label{fig:sen_pos_dist}
\end{figure}

We observed that the ROUGE performance on the PubMed$_\text{trunc}$ dataset is significantly lower than that on the PubMed dataset, with a 16.87 drop in R-1 for the extractive oracle and a 6.23 drop in R-1 for MemSum,\textit{ indicating that the introduction section is not sufficient to generate summaries close to the ground truth (abstracts).}
Even so, our model still significantly outperformed MatchSum on PubMed$_\text{trunc}$, and we attribute this improvement to the fact that MatchSum truncates the introduction section further to 512 tokens because it needs to compute document embeddings using Bert. 
Consequently, MatchSum extracts sentences mainly from the first 15 sentences of the document, while our MemSum produces a similar distribution of extracted sentence positions as the extractive oracle,
Figure \ref{fig:sen_pos_dist}. Thus, \textit{summarizing long documents is a non-trivial task}, and models that work well on summarizing short documents (e.g., CNN/DM) may fail to generalize to long documents.

\begin{table}
\footnotesize
% \tinysize
\centering
% \resizebox{0.5\textwidth}{!}{ 
\begin{tabularx}{\linewidth}{X}
\toprule 
\textbf{Human-written Summary:}

(...) While CMS is generally required to disallow, or \textit{\textbf{\textcolor{blue!100}{recoup, federal funds}}} from states for \textit{\textbf{\textcolor{blue!100}{eligibility-related improper payments}}} if the state's \textit{\textbf{\textcolor{blue!100}{eligibility error rate exceeds 3 percent}}}, it has not done so for decades, (...)
% because the method it used for calculating eligibility error rates was found to be insufficient for that purpose. To address this,
% in July 2017, 
CMS  \textit{\textbf{\textcolor{blue!100}{issued revised procedures through which it can recoup funds for eligibility errors, beginning in fiscal year 2022}}}. (...)\\
\hline
\textbf{Hepos-Sinkhorn (abstractive):}

(...)
The selected states also reported that they did not have adequate processes to address these issues. CMS has taken
steps to improve its oversight of the Medicaid program, including issuing guidance to states on the use of MAGI-exempt bases for determining eligibility, but these efforts have not been fully implemented. (...)\\
\hline
\textbf{MemSum (ours, extractive):}

(...)
% In 1983, CMS 
implemented its statutory requirement to \textit{\textbf{\textcolor{blue!100}{recoup funds}}}  associated with Medicaid \textit{\textbf{\textcolor{blue!100}{eligibility-related improper payments}}} for states  with an \textit{\textbf{\textcolor{blue!100}{eligibility error rate above 3 percent}}} through its MEQC program. (...)
% Because CMS  has  \textit{\textbf{not had a complete}} national estimate of improper payments due to eligibility errors since 2014, policymakers and other stakeholders have had an incomplete  picture of the extent of eligibility errors in the Medicaid program nationally. (...)
However, the agency has \textit{\textbf{\textcolor{blue!100}{introduced new procedures through which it  can, under certain circumstances, begin to recoup funds based on  eligibility errors in fiscal year 2022}}}. (...)
\\
\bottomrule
\end{tabularx}
% }
\caption{ \label{tab:res_gov-report-sample}
Comparison of the summary extracted by MemSum and the summary abstractively generated by Hepos-Sinkhorn \cite{huang2021efficient}. Compared with the abstractive summary, the MemSum-extracted summary has higher overlap with the human-written summary.} 
\end{table}

MemSum also significantly outperformed the state-of-the-art abstractive summarization model Hepos as measured by ROUGE scores, especially on the GovReport dataset.
A comparison of an exemplary MemSum-extracted summary and the corresponding Hepos-Sinkhorn-generated summary from the GovReport dataset (Table \ref{tab:res_gov-report-sample}) is consistent with the ROUGE comparison, showing that the MemSum-extracted summary is more accurate than the Hepos-Sinkhorn-generated summary and has higher overlap with the gold summary. 
We deem that this particularly good extraction performance on the GovReport dataset results from the higher ``extractiveness'' of the gold summaries in the GovReport dataset compared to other datasets, which may be due in part to technical language being difficult to abstractively summarize without a change in meaning. This is evidenced by the fact that the ROUGE scores of the extractive oracle on the GovReport dataset (Table \ref{tab:res_gov-report}) are higher than those of the PubMed and arXiv datasets (Table \ref{tab:res_pubmed}).
Therefore, \textit{extractive summarization may be more proper than abstractive summarization due to the requirement of stringent faithfulness of government report summaries}.

\subsection{Ablation Test}

\begin{table}
\centering
\resizebox{.8\linewidth}{!}{ 
\begin{tabular}{lccc}
\toprule
\multirow{1}*{\textbf{Model}}  & \multirow{1}*{\textbf{R-1}}  &   \multirow{1}*{\textbf{R-2}}   &  \multirow{1}*{\textbf{R-L}} 
% &  \multirow{2}*{\shortstack{\textbf{summ. len}\\\textbf{(\# sent.)}}}
 \\  \midrule
\textbf{MemSum} & \textbf{49.25} & \textbf{22.94} & \textbf{44.42} \\ %& 6.0 \\
MemSum w/o LSE & 48.12 & 22.04 & 43.36 \\%& 6.1  \\
MemSum w/o GCE & 46.85 & 20.31 & 41.95 \\ %& 6.0 \\
MemSum w/o EHE & 48.08 & 22.77 & 43.55 \\%& 7.0 \\ 
MemSum with GRU-EHE & 49.11 & 22.86 & 44.28 \\%& \\
MemSum w/o auto-stop & 48.25 & 22.63 & 43.70\\% & 7.0 \\
MemSum with ``STOP'' &  47.18 & 21.81& 42.20 \\%& 3.9 \\

\bottomrule
\end{tabular}
}
\caption{ \label{tab:res_ablation} Ablation study on the PubMed dataset. } 

\end{table}
% In order to analyze the individual contribution of different modules such as LSE, GCE and EHE, and the influence of different policies, automatic stopping mechanisms and module architectures, 
We conduct ablation studies by comparing the full MemSum model with the following \textit{variations in structures}: 1) MemSum w/o LSE, where we obtain local sentence embeddings by replacing the bi-LSTM based LSE by simple averages of word embeddings; 2) MemSum w/o GCE where we remove the GCE; 3) MemSum w/o EHE where we remove EHE, compute the scores for all sentences in one step, and samples sentences following the BanditSum policy \cite{dong2018banditsum}; 4) MemSum with GRU-EHE where we use a GRU to encode previously extracted sentences at each time step, and uses the last hidden state as the extraction history embedding for all remaining sentences, following  \citet{zhou-etal-2018-neural-document}.  

Meanwhile, we also tested two variations that adopted \textit{different stopping mechanisms}:  1) MemSum w/o auto-stop that does not stop extraction automatically based on $p_\text{stop}$, but that extracts a fixed number of sentences; 2) MemSum with ``STOP'' that inserts a  special  stop sentence  (e.g.   “STOP")  into  the document, and stops extraction once the agent selects this sentence.

\noindent\textbf{Contribution of Modules.} 
Removing GCE has a greater impact on performance than removing LSE (Table \ref{tab:res_ablation}), suggesting that modeling global contextual information is more critical than modeling local sentence information in our MemSum framework, which contrasts with the result that modeling local sentence information is more important in the Atten-Cont \cite{xiao-carenini-2019-extractive} framework. Furthermore, we observed a significant performance degradation when removing EHE, but no significant difference between MemSum and MemSum with GRU-EHE, indicating that EHE is necessary, but our \textit{MemSum policy is not strongly dependent on the specific structure of this module} (e.g., attention-based or RNN-based). 

\noindent\textbf{Influence of Stopping Mechanisms.} MemSum w/o auto-stop achieves lower ROUGE scores than MemSum, revealing the necessity of auto stopping in our MemSum architecture. Meanwhile, MemSum with ``STOP'' produced summaries with fewer extracted sentences (3.9 vs. 6.0 sentences on average) and significantly lower ROUGE scores. We attribute this reduction to the predictable positive reward obtained from selecting the special stop sentence that ends an episode, which leads to a preference for this final action and increases the likelihood of taking this action prematurely.

\subsection{History Awareness Avoids Redundancy}

\begin{table}
\centering
\resizebox{\linewidth}{!}{ 
\begin{tabular}{lcccc}
\toprule
\multirow{2}*{\textbf{Model}}  & \multirow{2}*{\textbf{R-1}}  &   \multirow{2}*{\textbf{R-2}}   &  \multirow{2}*{\textbf{R-L}}  &   \multirow{2}*{\shortstack{\textbf{ duplicate}\\\textbf{percentage} }}
 \\ \\  \midrule
\textbf{MemSum} & \textbf{49.16} & \textbf{22.78} & \textbf{44.39} &  0\%\\
MemSum w/o auto-stop & 48.21 & 22.59 & 43.76 & 0\% \\
MemSum w/o EHE & 42.82 & 18.18 & 36.68 &  41\% \\  %2.9 \\ 
\multirow{2}*{\shortstack{MemSum w/o EHE$\ \ \ \ \ \ \ $ \\ $\ \ \ \ \ \ \ \ \ \ \ $+3gram blocking}}& \multirow{2}*{46.85} & \multirow{2}*{19.93} & \multirow{2}*{42.40} & \multirow{2}*{0\%} \\ \\

\bottomrule
\end{tabular}
}
\caption{ \label{tab:res_ablation_avoid_redundancy} Performance on the redundant PubMed dataset.}  
% The history-aware MemSum$_{full}$ model is much less affected by the redundancy introduced in the document than the model without history awareness. } 
\end{table}

We hypothesized that the extraction history allows MemSum to avoid sentences that are similar to existing sentences in the current partial summary, intuitively mimicking what humans do when extractively summarizing documents. To verify this, we created a redundant PubMed dataset in which we repeated each sentence in the document, with the replicated sentences immediately following the originals. On this dataset, we trained and tested MemSum and MemSum w/o EHE (no history awareness), and we compared different models in terms of ROUGE scores and average \textit{duplicate percentage} that is defined as the average percentage of the duplicated sentences among all extracted sentences in a summary.

As reported in Table \ref{tab:res_ablation_avoid_redundancy}, for  MemSum w/o EHE, on average $41\%$ of sentences in the extracted summaries were duplicated. Along with the high duplicate ratio came a significant decrease in ROUGE score. By contrast, the performance of the full MemSum model with history awareness was only slighted affected when comparing the results of the MemSum on the PubMed dataset (Table \ref{tab:res_pubmed}) and on the redundant PubMed dataset (Table \ref{tab:res_ablation_avoid_redundancy}).

Meanwhile, using the Trigram Blocking method that skips a sentence if it has a trigram that overlaps with the current summary \cite{liu2019text} is also successful in avoiding repetitive sentences. However, the ROUGE scores associated with Trigram Blocking were significantly lower than those of the  MemSum with awareness of extraction history. In summary, the history-aware MemSum model spontaneously learns an optimized strategy to avoid redundant sentences without explicit human guidance or crude rules, and thus shows better performance.

\begin{figure}
\centering
  \includegraphics[width=\linewidth]{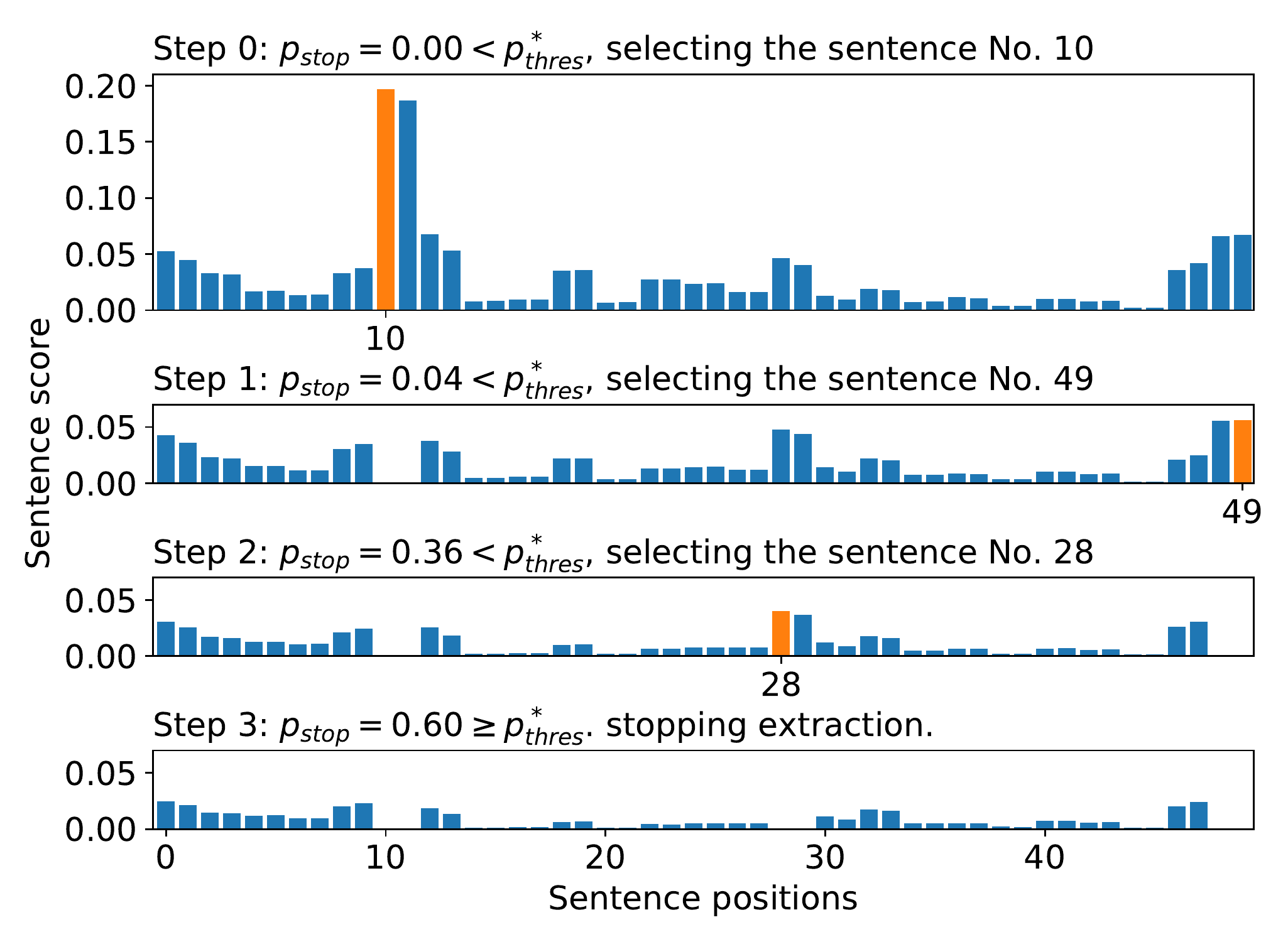}
  \caption{The sentence scores of 50 sentences computed by MemSum at extraction steps 0 to 3. In the document, there is artificial redundancy in that the $(2n)_{\text{th}}$ and the $(2n+1)_{\text{th}}$ sentences are identical ($n=0,1,...,24$). }
  \label{fig:score_change}
\end{figure}

\noindent\textbf{Case Study: How does MemSum Avoid Redundancy?} 

We let MemSum summarize a document sampled from the test set of the redundant PubMed dataset and monitored the sentence scores produced by the Extractor during each extraction step. The results are shown in Figure \ref{fig:score_change}. At time step $0$, the $10_{\text{th}}$ sentence obtained the maximum score and was thus selected into the summary. At time step $1$, we noticed that the $11_\text{th}$ sentence, which is a replica of the $10_\text{th}$ sentence, had a score close to zero. The same was also true for the other selected sentences and their following sentences, revealing competent repetition avoidance of the Extractor. 
%Thus, we see that knowing the extraction history helps MemSum to avoid selecting redundant sentences. Once a sentence is selected for the summary, the Extractor of the full MemSum produces a very low score for the duplicate sentence in the following time steps, so that this duplicate sentence is rarely selected. 
Because the EHE is insensitive to the extraction order and to sentence position information, as described in Section \ref{sec:policy_network_structure}, we can conclude that the full MemSum avoids redundancy by evaluating the similarity between selected and remaining sentences, rather than by ``remembering" selected sentences' positions.

% the position of the previously selected sentences.

\subsection{Human Evaluation}
\label{sec:human-evaluation}
We conducted human evaluation following \citet{DBLP:journals/corr/abs-1804-07036,dong2018banditsum,luo2019reading}.
For each document sampled from the test set of the PubMed dataset, we provide a reference summary, and volunteers are asked to rank a pair of randomly ordered summaries produced by two models according to three criteria: non-redundancy, coverage, and overall quality. The better model will be ranked \#1 while the other is ranked \#2, and if both models extract the same summary, then they will both get the \#1 rank. In experiment 1, we compared NeuSum, which always extracts $7$ sentences, and MemSum, which extracts a flexible number of sentences thanks to automatic stopping. In experiment 2, we discounted for differences in the number of extracted sentences by making MemSum w/o auto-stop to also extract $7$ sentences.
A user-friendly interactive web interface was implemented to assist the evaluation process, with details in Appendix~\ref{sec:web-interface}.

% \begin{table}
% \centering
% \resizebox{\linewidth}{!}{ 
% \begin{tabular}{clcccc}
% \toprule
% \multirow{2}*{\shortstack{\textbf{Experi-}\\\textbf{ment} }  } &
% \multirow{2}*{\textbf{Model}}  & \multirow{2}*{\textbf{overall}}  &   \multirow{2}*{\textbf{coverage}}   &  \multirow{2}*{\shortstack{\textbf{non-}\\\textbf{redundancy}}}  & \multirow{2}*{\textbf{Avg. summ. len.}} 

%  \\ \\  \midrule
% \multirow{2}*{I}   & NeuSum & 1.58 & \textbf{1.46} & 1.68 \\
%  & MemSum & \textbf{1.37} & 1.49  & \textbf{1.28}  \\ \midrule
% \multirow{3}*{II}   &  NeuSum & 1.57 & \textbf{1.44} & 1.65 \\
%  & MemSum & \multirow{2}*{\textbf{1.38}} & \multirow{2}*{1.51}  & \multirow{2}*{\textbf{1.30}}  \\
%  & $\ \ \ \ $w/o auto-stop
%  \\
% \bottomrule
% \end{tabular}
% }
% \caption{ \label{tab:res_human_evaluation} The average ranking of NeuSum and MemSum is reported. The smaller the ranking, the better the model. Four volunteers participated in these experiments, and evaluated 65 and 63 pairs of summaries in Experiment 1 and 2, respectively. } 

% \end{table}

\begin{table}
\centering
\resizebox{\linewidth}{!}{ 
\begin{tabular}{lcccc}
\toprule
\multirow{3}*{Criteria} & \multicolumn{2}{c}{Experiment I} &\multicolumn{2}{c}{Experiment II} \\
\cmidrule(lr){2-3}
\cmidrule(lr){4-5}

 & \multirow{2}*{NeuSum} & \multirow{2}*{MemSum} & \multirow{2}*{NeuSum} & \multirow{2}*{\shortstack{MemSum w/o\\{auto-stop}}} \\
 \\
 \midrule
 overall & 1.58 & \textbf{1.37} & 1.57 & \textbf{1.38} \\
 coverage & \textbf{1.46} & 1.49 & \textbf{1.44} & 1.51  \\
 non-redundancy & 1.67 & \textbf{1.28}* & 1.65 & \textbf{1.30}* \\
avg. summ. length\\
$\ \ \ \ \ \ \ $\# of sentences&  7.0 &  \textbf{5.6}* & 7.0 & 7.0\\
$\ \ \ \ \ \ \ $\# of words& 248.8 & \textbf{189.3}* & 263.6 & \textbf{239.5}*
\\
 
 \bottomrule

\end{tabular}
}
\caption{ \label{tab:res_human_evaluation} The average ranking of NeuSum and MemSum is reported. The smaller the ranking, the better the model. Four volunteers participated in these experiments, and evaluated 67 and 63 pairs of summaries in Experiment 1 and 2, respectively. ``*'' indicates statistical significance (p<0.005) in a Wilcoxon signed-rank test \cite{Woolson2007}. }

\end{table}

Table \ref{tab:res_human_evaluation} reports the human evaluation results for both experiments. Both MemSum and MemSum w/o auto-stop ranked significantly higher (p<0.005) than NeuSum in terms of non-redundancy and achieved a better average overall quality. In terms of word count, MemSum produces shorter summaries than NeuSum in both experiments, even though both models extract the same number of sentences in experiment 2. These results show that redundancy avoidance of MemSum is particularly good, even without the auto-stop mechanism. 
The slightly better performance of NeuSum in terms of coverage needs to be weighed against it extracting significantly longer summaries. Note that neither NeuSum nor our model is trained to optimize the order of the extracted sentences. 
Therefore, we did not use fluency, which depends on sentence order, as a metric for human evaluation. Improving the fluency of the extracted summaries will be the subject of our future research.

\section{Conclusion}

Extractive summarization can be achieved effectively with a multi-step episodic Markov decision process with history awareness. Using encoders of local sentence, global context, and extraction history,  MemSum is given information that is intuitively also used by humans when they summarize a document. Awareness of the extraction history helps MemSum to produce compact summaries and to be robust against redundancy in the document. As a lightweight model (Appendix \ref{sec:ref_sum_time}), MemSum outperforms both extractive and abstractive baselines on diverse long document summarization tasks. Because MemSum achieves SOTA performance on these tasks, MDP approaches will be promising design choices for further research.

\section*{Acknowledgements}
We acknowledge support from the Swiss National
Science Foundation (grant 31003A\_182638) and the NCCR Evolving Language, Swiss National Science Foundation Agreement No. 51NF40\_180888. We also thank the anonymous reviewers for their useful comments.

% % Entries for the entire Anthology, followed by custom entries
\bibliography{anthology,custom}
\bibliographystyle{acl_natbib}

\appendix

\section{Computing Hardware}
We trained our \textsc{MemSum} model and its variations on 8 NVIDIA GeForce RTX 2080 Ti 11GB GPUs. During testing, we used a single NVIDIA TITAN X Pascal 12GB GPU.

\section{Comparison of Validating and Testing Performance }

We compare the validating and testing performance of the MemSum model on the following datasets: PubMed \cite{cohan2018discourse}, arXiv \cite{cohan2018discourse}, and GovReport \cite{huang2021efficient}. The results are reported in Table \ref{tab:val_test_comparison}.

\section{Summarization Time}
\label{sec:ref_sum_time}
We analyzed the average time taken by MemSum to extractively summarize a source document from the test set. The average summarizaion time is positively correlated with the document length and the number of extracted sentences, Table \ref{tab:extraction_time_analysis}. On the one hand, on longer documents, it takes longer to compute the scores of remaining sentences, which delays the action of either stopping extraction or selecting a sentence. On the other hand, the more sentences must be extracted, the more actions are needed of selecting sentences within an episode. 
\begin{table}
\setlength\tabcolsep{3pt} 
\centering
\resizebox{\linewidth}{!}{ 
\begin{tabular}{lcccccccc}
\toprule
\multirow{2}*{\textbf{Datasets}} & & \multicolumn{3}{c}{\textbf{Validating}}  &  & \multicolumn{3}{c}{\textbf{Test}}
 \\ % \cline{2-3} \cline{6-8} 
 && \textbf{R-1} & \textbf{R-2} & \textbf{R-L} && \textbf{R-1}&\textbf{R-2} &\textbf{R-L} \\ 
 \midrule
PubMed && 49.14 & 22.92 & 44.33 && 49.25 & 22.94 & 44.42 \\ 
arXiv&& 48.23 & 20.17 & 42.31 && 48.42 & 20.30 & 42.54 \\ 
PubMed$_\text{trunc}$ && 43.46 & 16.77 & 38.65 && 43.08 & 16.71 & 38.30 \\ 
GovReport && 59.29& 28.57 & 56.46 && 59.43 & 28.60 & 56.69 \\ 
\bottomrule
\end{tabular}
}
\caption{ \label{tab:val_test_comparison} Validating and testing scores of the MemSum model tested on the PubMed, the arXiv and the GovReport datasets. }
\end{table}

\begin{table}
\centering
\resizebox{\linewidth}{!}{ 
\begin{tabular}{lccc}
\toprule  
\multirow{3}*{\textbf{Datasets}} & \multirow{3}*{ \shortstack{ \textbf{avg. doc.}\\\textbf{length} \\ \textbf{(words)}} } &\multirow{3}*{\shortstack{\textbf{Avg. extractive }\\\textbf{summ. length}\\\textbf{(\# sentences)}}}&\multirow{3}*{\shortstack{\textbf{Avg. extractive}\\\textbf{summ. time}\\\textbf{(ms)}}}\\ 
\\ 
\\
 \midrule
PubMed & 2,730 & 6.0 $\pm$ 1.2 & 91.7 $\pm$ 8.6 \\
arXiv & 5,206 & 4.8$\pm$ 0.5 &  114.0 $\pm$ 5.0 \\
PubMed$_\text{trunc}$ & 408 & 5.3$\pm$ 1.4 &  27.7 $\pm$ 4.6 \\
GovReport &  7,932 &  21.7 $\pm$ 1.8 & 197.0 $\pm$ 14.8  \\

\bottomrule
\end{tabular}
}
\caption{ \label{tab:extraction_time_analysis} Average extractive summarization time of  MemSum on different datasets.}
\end{table}

\section{Selection of optimal stopping threshold}
\label{sec:selection_of_optimal}
\begin{figure}[ht]
\centering
  \includegraphics[width=\linewidth]{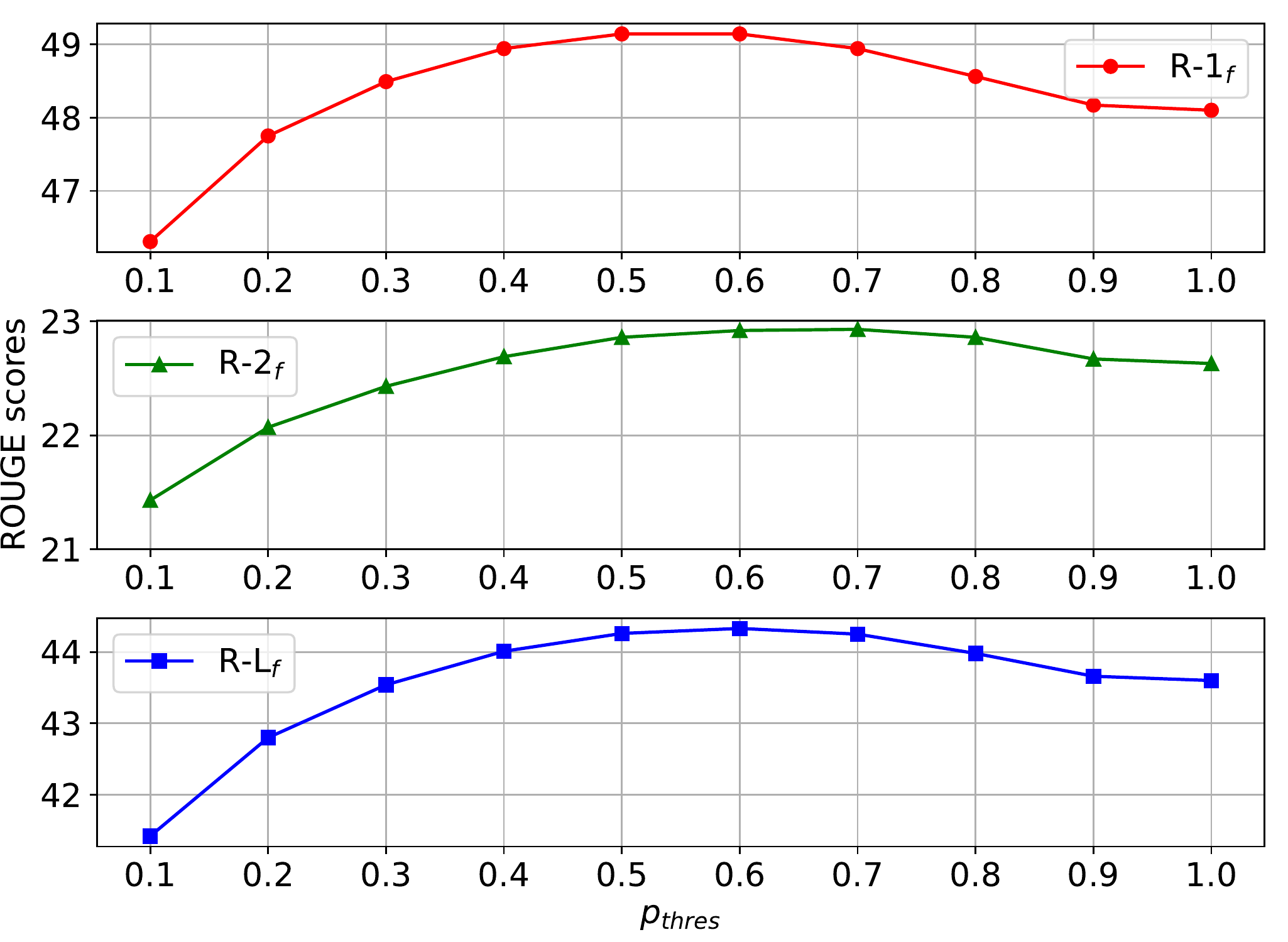}
  \caption{The ROUGE scores for different stopping thresholds $p_\text{thres}$ on the PubMed validating set.}
  \label{fig:pthres_pubmed}
\end{figure}

\begin{figure}[ht]
\centering
  \includegraphics[width=\linewidth]{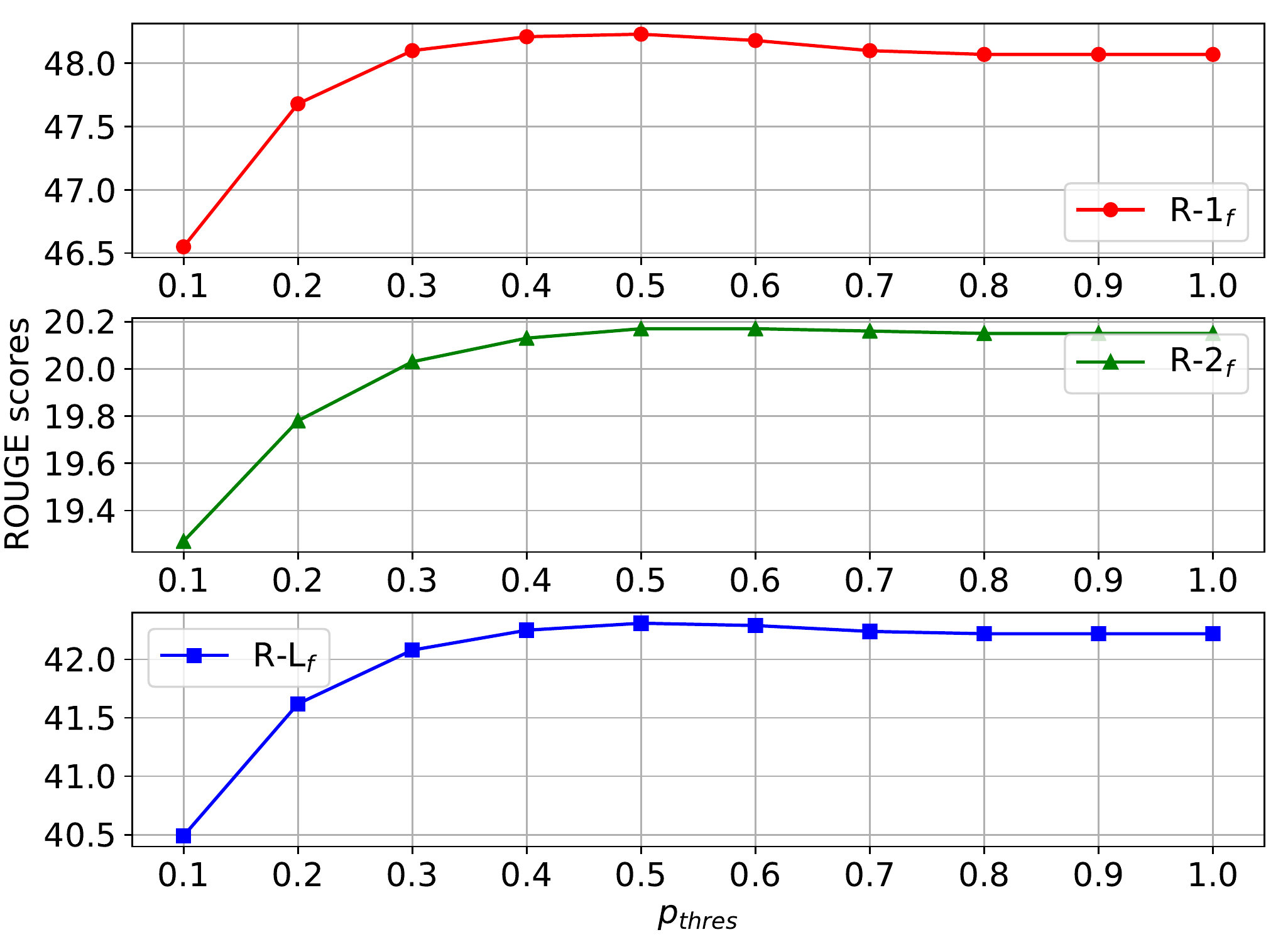}
  \caption{The ROUGE scores for different stopping thresholds $p_\text{thres}$ on the arXiv validating set.}
  \label{fig:pthres_arxiv}
\end{figure}
\begin{figure}[ht]
\centering
  \includegraphics[width=\linewidth]{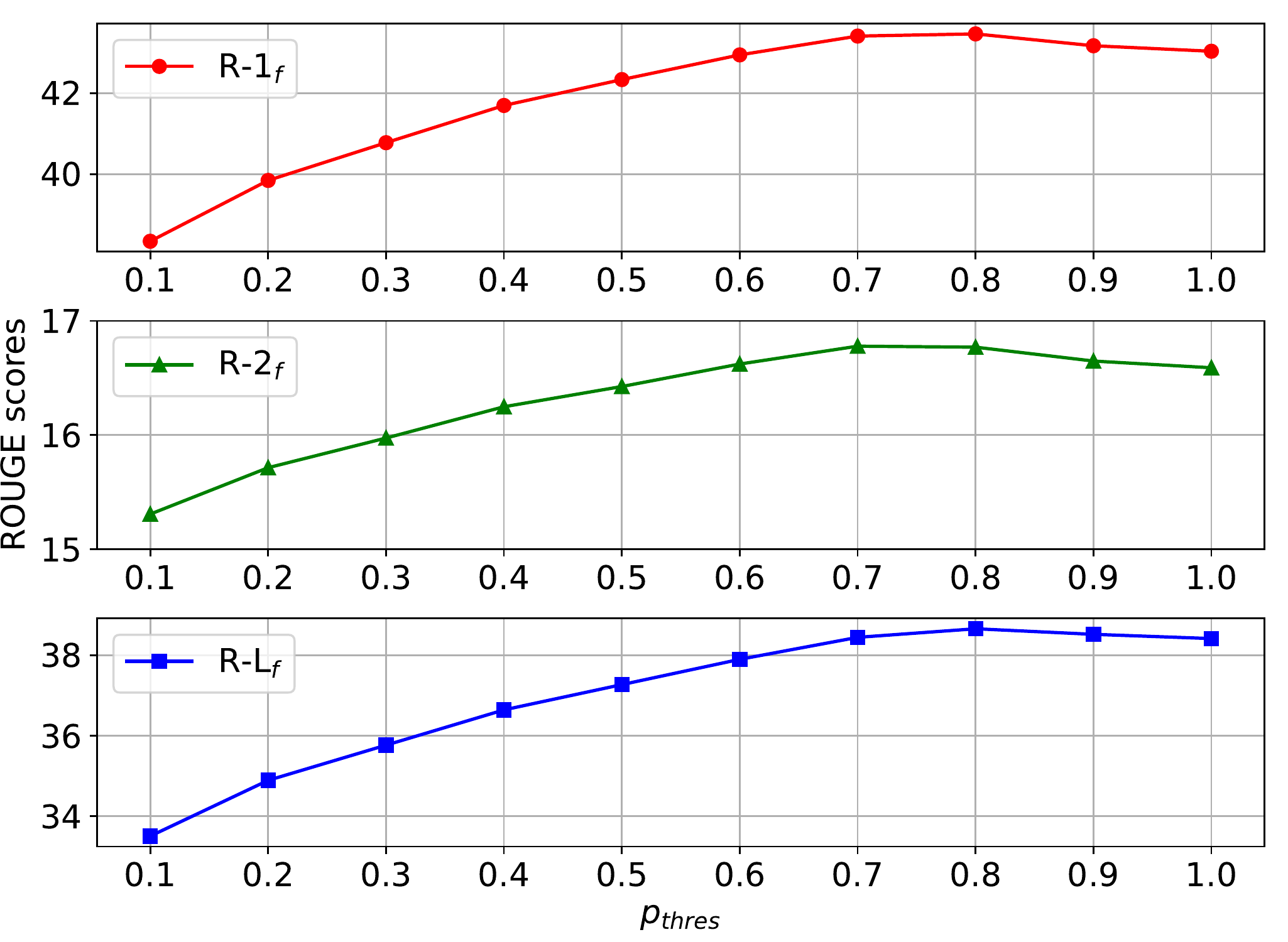}
  \caption{The ROUGE scores for different stopping thresholds $p_\text{thres}$ on the PubMed$_\text{trunc}$ validating set.}
  \label{fig:pthres_pubmed_trunc}
\end{figure}

The stopping threshold $p_{\text{thres}}$ is an important hyperparameter that sets the stopping probability in an episode, as described in the Implementation Details. We determined the optimal stopping threshold $p^*_{\text{thres}}$ as follows: For each data set and each stopping threshold $p_{\text{thres}}\in\{ 0.1, 0.2,\dots,1.0\}$, we chose as optimal stopping threshold $p^*_{\text{thres}}$ the one with maximal ROUGE score on the corresponding validating set. 

\begin{figure}[ht]
\centering
  \includegraphics[width=\linewidth]{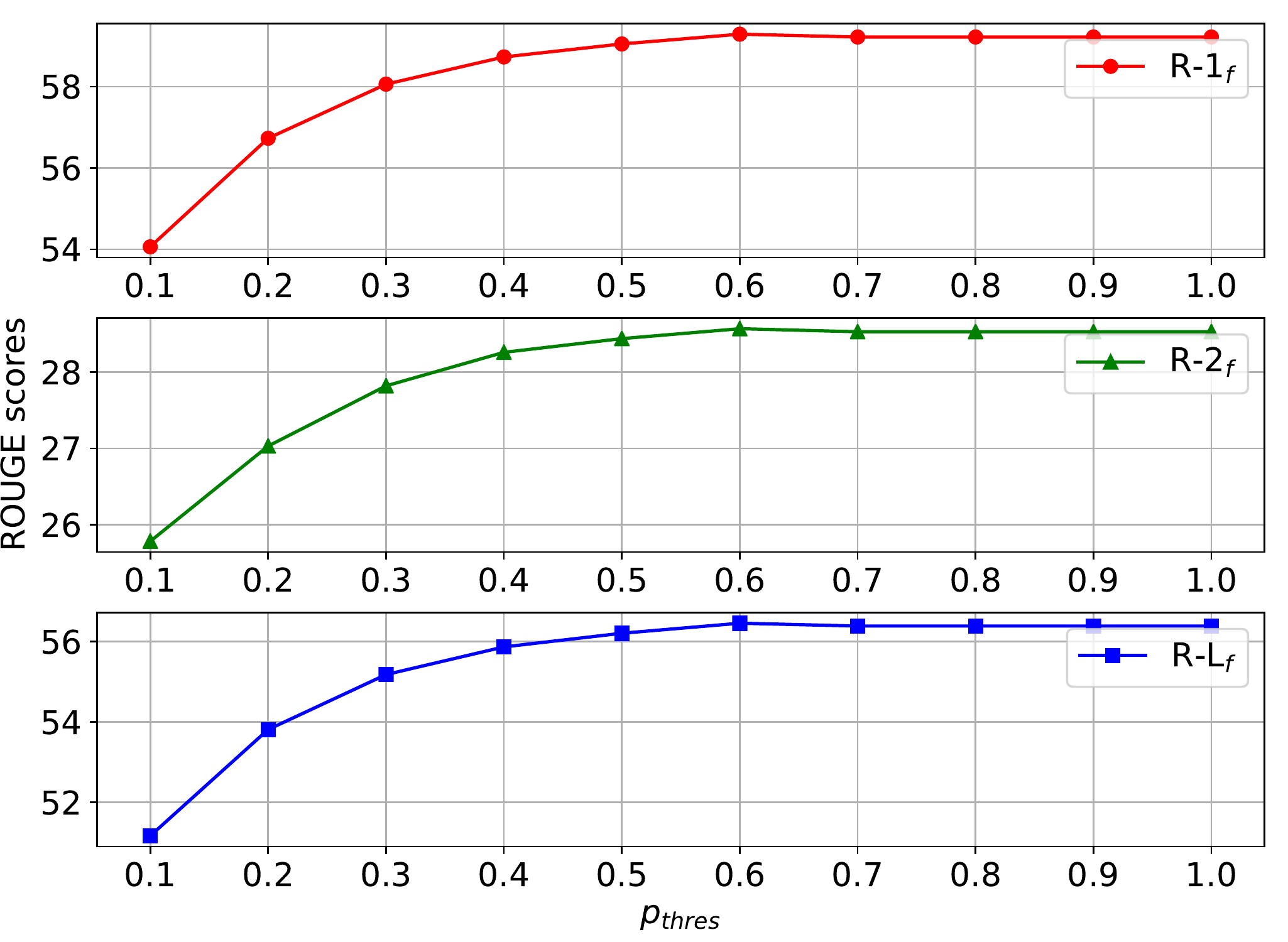}
  \caption{The ROUGE scores for different stopping thresholds $p_\text{thres}$ on the GovReport validating set.}
  \label{fig:pthres_govreport}
\end{figure}

The ROUGE scores as a function of stopping threshold are shown in Figure \ref{fig:pthres_pubmed}, \ref{fig:pthres_arxiv} and \ref{fig:pthres_govreport} on the validating set of the PubMed, the arXiv, and the GovReport data set, respectively. The functions exhibit a local maximum between $0.1$ and $1.0$, which implies that when $p_\text{thres}$ is too low, summaries tend to be too short, while when $p_\text{thres}$ is too high, summaries will be unduly lengthy. We chose $p^*_{\text{thres}}=0.6,\ 0.5,\ 0.8\ \text{and}\ 0.6$ for the PubMed, the arXiv, the PubMed$_\text{trunc}$, and the GovReport dataset, respectively.

\section{Creating High-ROUGE Episodes for Training}
\label{sec:high_rouge}
As introduced in Section $3.4$ and Algorithm $1$ in the main paper, at each training iteration, we sampled a high-ROUGE episode from the set $\mathbb{E}_p$. An episode can be viewed as a sequence of state-action pairs as well as the final reward, such as  ($S_0$,$s_{a_0}$,$\dots$,$S_{T-1}$,$s_{a_{T-1}}$, $S_{T}$,$A_{\text{stop}}$,$r$). Here, $\{s_{a_0}\dots  s_{a_{T-1}}\}$ is the extracted summary consisting of a set of $T$ sentences, and $r$ is the average of the associated ROUGE-1, ROUGE-2, and ROUGE-L F1 scores.

In \cite{nallapati2016summarunner}, a greedy approach was proposed to select candidate summaries by sequentially selecting from the source document the optimal sentence that maximally improves the average ROUGE-1/2/L score once added to the current subset of selected sentences.

In this paper, we define a high-ROUGE episodes set $\mathbb{E}_p$ as the set of multiple episodes where each episode has a high average ROUGE-1/2/L F1 score. To obtain not a single episode in $\mathbb{E}_p$ but multiple episodes with high average ROUGE-1/2 scores, we modified the greedy approach by considering not only the optimal sentence at each sentence selection step but also $B-1$ sub-optimal sentences. This sentence-sampling step is repeated for each of these $B$ new subsets to result in a potentially exponentially growing number of high ROUGE-score episodes. This process stops until no sentence can further improve the average ROUGE-1/2/L score or a maximum number $N_\text{max}$ of selected sentences per episode is reached. $B$ can be considered the branching size, analogous to beam search strategies in neural machine translation \cite{sutskever2014sequence,freitag2017beam}. We set $B=2$ by default.

In practice, we notice that ROUGE-L F1 score is computationally intensive. Because when creating $\mathbb{E}_p$ we need to iteratively re-compute ROUGE scores once a new sentence is added to the current summary, including the ROUGE-L F1 score into computation would heavily slow down the process of creating the high-ROUGE episodes set for training. 
As a compromise, we do not incorporate the ROUGE-L F1 score into the intermediate steps of our modified greedy approach. Instead, we calculate the ROUGE-L F1 score only once after a complete high-ROUGE episode is selected, and use this ROUGE-L F1 score together with ROUGE-1/2 F1 scores to compute the reward $r$ for each episode. A similar strategy was adopted in \citet{zhou-etal-2018-neural-document} to create the training dataset by maximizing ROUGE-2 F1 scores only.

We refer to an episode  ($S_0$,$s_{A}$,$S_{1}$,$s_{B}$,$S_{2}$,$s_{C}$,$S_{3}$,$A_{\text{stop}}$,$r$) as ``($s_A,s_B,s_C$)" for simplicity. Because permuted episodes ($s_A,s_B,s_C$), ($s_A,s_C,s_B$), and ($s_C,s_B,s_A$) have nearly the same average ROUGE-1/2 scores (although ROUGE-L score may differ), we decided to equally sample them with the hope to avoid overfitting. This decision does not interfere with our usage of extraction history, because under ($s_A,s_B,s_C$), the agent learns to extract $s_C$ from $\{s_A,s_B\}$, while under ($s_C,s_B,s_A$) it learns to extract $s_A$ from $\{s_B,s_C\}$. Thus, history plays a role in both cases.

\section{Padding and Truncation of Sentences and Documents}
In the training process, we used mini-batch gradient descent. To enable efficient batch-wise parallel GPU computation, each document in a mini batch needs to have the same number of sentences, and each sentence needs to have the same number of tokens. Therefore, in order to unify the sentence length to a common value $L_\text{sen}$, we appended ``PAD" tokens at the end of sentences shorter than $L_\text{sen}$, and we truncated sentences longer than $L_\text{sen}$. To  unify the document length in terms of number of sentences to a common value $L_\text{doc}$, we appended empty-string sentences at the end of documents shorter than $L_\text{doc}$, and truncated documents longer than $L_\text{doc}$. To ensure consistency between training and testing we also performed the same padding and truncation setting during testing. We set $L_\text{doc}$ to $500$ for the PubMed, the arXiv, and the GovReport datasets and $50$ for the PubMed$_\text{trunc}$ dataset based on the document length statistics shown in Table \ref{tab:datasets} in the main paper. We set $L_\text{sen}$ to $100$ for the PubMed, the PubMed$_\text{trunc}$, and the GovReport datasets and $150$ for the arXiv dataset, because we noticed a larger variance in the length of sentences in the arXiv dataset.

\section{Interactive Web Interface for Human Evaluation}
\label{sec:web-interface}
\begin{figure*}
\centering
  \includegraphics[width=\linewidth]{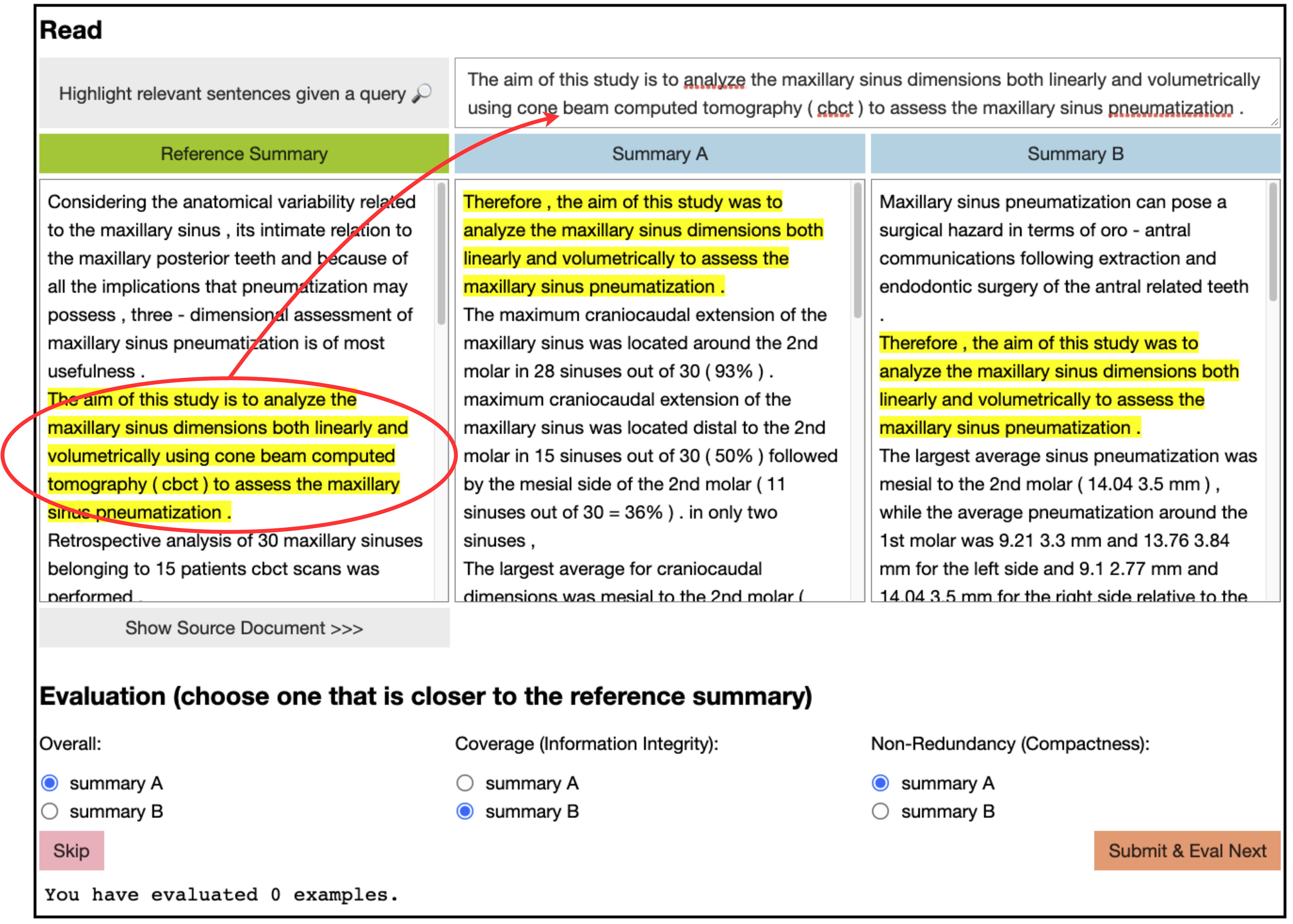}
  \caption{We designed an interactive web interface for the human evaluation experiments introduced in Section \ref{sec:human-evaluation}.}
  \label{fig:web_interface}
\end{figure*}

To provide for a convenient evaluation procedure for volunteers, we designed an interactive web interface based on Jupyter Widgets\footnote{\url{https://ipywidgets.readthedocs.io/}}. As shown in Figure \ref{fig:web_interface}, for each document, we display the reference summary, summary A, and summary B from left to right. The reference summary contains the ground-truth abstract. Summaries A and B are the summaries extracted by the two models assigned in a random order, so that the volunteers do not know which model either summary came from. Meanwhile, the volunteers were allowed to read the source document by clicking the button ``Show Source Document $>>>$''. We also provided a sentence highlighting function to help the volunteers rapidly retrieve relevant content. We allowed evaluators to copy a sentence from the reference summary and paste it to the text box above. After clicking the button ``Highlight relevant sentences given a query'', relevant sentences in both summaries were highlighted, to help the volunteers rapidly find information of interest. The relevance score of a pair of sentences was given by the cosine similarity of the two sentences' embeddings computed with Sent2vec \cite{pgj2017unsup}. In the evaluation panel the volunteers selected the better summary (A or B) by comparing the model-produced summary with the reference summary on three criteria: overall quality, coverage (in terms of information content), and non-redundancy. After making a choice they clicked the button ``Submit \& Eval Next" to submit the current evaluation result and evaluate the next summaries, or click ``Skip" if they were not sure which summary was indeed better.

\section{Examples of Extracted Summaries}

\begin{table*}[ht]
\centering
% \resizebox{\linewidth}{!}{ 
\begin{tabularx}{\linewidth}{l|X}
\toprule 
\textbf{Title} & BERT: Pre-training of Deep Bidirectional Transformers for Language Understanding\\ \midrule
\shortstack{\textbf{Original}\\\textbf{abstract}} & \colorbox{yellow}{We introduce a new language representation model called BERT, which stands for} \colorbox{yellow}{Bidirectional Encoder Representations from Transformers.} \colorbox{cyan}{Unlike recent langua-} \colorbox{cyan}{ge representation models (Peters et al., 2018a; Radford et al., 2018), BERT is } \colorbox{cyan}{designed to pretrain deep bidirectional representations} from unlabeled text by jointly conditioning on both left and right context in all layers. As a result, the pre-trained BERT model can be finetuned with just one additional output layer to create state-of-the-art models for a wide range of tasks, such as question answering and language inference, without substantial taskspecific architecture modifications. BERT is conceptually simple and empirically powerful. \colorbox{pink}{It obtains new state-of-the-art} \colorbox{pink}{results on eleven natural language processing tasks,} including pushing the GLUE score to 80.5\% (7.7\% point absolute improvement), MultiNLI accuracy to 86.7\% (4.6\% absolute improvement), SQuAD v1.1 question answering Test F1 to 93.2 (1.5 point absolute improvement) and SQuAD v2.0 Test F1 to 83.1 (5.1 point absolute improvement)\\ \midrule

\shortstack{\textbf{DANCER}\\ \textbf{PEGASUS }} & 
Language model pre-training has been shown to be effective for improving many natural language processing tasks such as sentence-level paraphrasing and entity recognition tasks. However, current approaches to pre-trained language models are restricted to unidirectional language models. \colorbox{yellow}{In this paper, we propose a new approach to pre-} \colorbox{yellow}{trained language models based on bidirectional encoder transformers (BERT).} BERT is inspired by the pre-training objective of cloze task (Taylor et al., 1953), where the goal is to predict some masked language representations from the input. We introduce BERT and its detailed implementation in this paper. The BERT model is first initialized with the pre-trained parameters, and all of the parameters are fine-tuned using labeled data from the downstream tasks. Rich unsupervised pre-training is an integral part of many language understanding systems. In particular, these results enable even low-resource tasks to benefit from deep unidirectional architectures. Our major contribution is further \colorbox{pink}{generalizing these findings to deep bidirectional} \colorbox{pink}{architectures, allowing the same pre-trained model to successfully tackle a broad set} \colorbox{pink}{of NLP tasks.} \\ \midrule
\textbf{ROUGE1-F1} & 36.52 \\ \midrule
\shortstack{\textbf{MemSum}} & Language model pre-training has been shown to be effective for improving many natural language processing tasks. \colorbox{yellow}{In this paper, we improve the fine-tuning based} \colorbox{yellow}{approaches by proposing BERT: Bidirectional Encoder Representations from Trans-} \colorbox{yellow}{formers.} The masked language model randomly masks some of the tokens from the input, and the objective is to predict the original vocabulary id of the masked word based only on its context. \colorbox{cyan}{Unlike Radford et al. (2018), which uses unidirectional} \colorbox{cyan}{language models for pre-training, BERT uses masked language models to enable} \colorbox{cyan}{pretrained deep bidirectional representations.} BERT is the \colorbox{pink}{first finetuning based} \colorbox{pink}{representation model that achieves state-of-the-art performance on a large suite of} \colorbox{pink}{sentence-level and token-level tasks,} outperforming many task-specific~ architectures. 
 \\ \midrule
\textbf{ROUGE1-F1} & 44.29 \\

\bottomrule
\end{tabularx}
% }
\caption{ \label{tab:summarization_example_bert} Example summaries for Dancer Pegasus \cite{dancerp} and MemSum. }
\end{table*}

\begin{table*}[ht]
\centering
% \resizebox{\linewidth}{!}{ 
\begin{tabularx}{\linewidth}{l|X}
\toprule 
\textbf{Title} & (This paper) MemSum: Extractive Summarization of Long Documents using Multi-step Episodic Markov Decision Processes\\ \midrule
\textbf{Original abstract} & \colorbox{yellow}{We introduce MemSum (Multi-step Episodic Markov decision process ex-} \colorbox{yellow}{tractive SUMmarizer), a reinforcement-learning-based extractive summarizer} enriched at each step with information on the current extraction history. When MemSum iteratively selects sentences into the summary, it considers a broad information set that would intuitively also be used by humans in this task: \colorbox{cyan}{1) the text content of the sentence, 2) the global text context of the rest of} \colorbox{cyan}{the document, and 3) the extraction history consisting of the set of sentences} \colorbox{cyan}{that have already been extracted.} With a lightweight architecture, MemSum \colorbox{pink}{obtains state-of-the-art test-set performance (ROUGE) in summarizing long} \colorbox{pink}{documents taken from PubMed, arXiv, and GovReport.} Ablation studies demonstrate the importance of local, global, and history information. A human evaluation confirms the high quality and low redundancy of the generated summaries, stemming from MemSum's awareness of extraction history.
\\ \midrule

\textbf{MemSum summary} & \colorbox{yellow}{In this paper, we propose to model extractive summarization as a multi-step} \colorbox{yellow}{episodic Markov Decision Process (MDP).}
As shown in Figure 1, at each time step in an episode, we define a sentence state composed of three sub-states: \colorbox{cyan}{1) the local content of the sentence, 2) the global context of} \colorbox{cyan}{the sentence within the document, and 3) information on the extraction} \colorbox{cyan}{history, including the previously selected set of unordered sentences and} \colorbox{cyan}{the remaining sentences.}
To efficiently encode local and global sentence states, we design an extraction agent based on LSTM networks.
We show that extraction-history awareness allows our model to extract more compact summaries than models without history awareness and behave more robustly to redundancies in documents. 3) \colorbox{pink}{Our model outperforms both extractive and} \colorbox{pink}{abstractive summarization models on PubMed, arXiv, and GovReport datasets.}
 \\ \midrule
\textbf{ROUGE1-F1} & 48.57 \\

\bottomrule
\end{tabularx}
% }
\caption{ \label{tab:summarization_example_memsum} MemSum summary of this paper. }
\end{table*}

We provide summarization examples in Table \ref{tab:summarization_example_bert} and \ref{tab:summarization_example_memsum}. In Table \ref{tab:summarization_example_bert}, we compared MemSum trained on the arXiv dataset with Dancer Pegasus \cite{dancerp} on a typical paper on which MemSum achieved higher ROUGE-1 F score than Dancer Pegasus. In Table \ref{tab:summarization_example_memsum} we provide the extractive summary of this paper itself using our MemSum model. Sentences with similar meanings in different summaries are highlighted in the same color.

\section{Reproducibility}
\label{sec:reproduce}
The MemSum code and variants of MemSum that we used in our ablation study, as well as the MemSum parameters trained on the PubMed dataset, can be found in the submitted code.zip file. Also, we provide a sample of the datasets used in this paper in the data.zip file, as well as the raw data for the human evaluation. This will ensure that the results in this work are well reproducible.

\end{document}